\documentclass[runningheads]{llncs}
\usepackage[T1]{fontenc}
\usepackage{graphicx}
\usepackage{amsmath,amssymb,amsfonts}
\usepackage{algorithm}
\usepackage{algorithmicx}
\usepackage{algpseudocode}
\usepackage{booktabs}
\usepackage{multirow}
\usepackage{subcaption}
\usepackage{array}
\usepackage{comment}
\usepackage{enumerate}

\newtheorem{myexample}{Example}

\floatname{algorithm}{Procedure}

\newcommand{\INTCODE}{\mathtt{InteractCode}}
\newcommand{\INTSTRUC}{\mathtt{InteractStruc}}
\newcommand{\IC}{\mathtt{iCode}}
\newcommand{\IS}{\mathtt{iStruc}}
\newcommand{\IP}{\mathtt{iProg}}

\newcommand{\Ratify}{{\mathit{RATIFY}}}
\newcommand{\Revise}{{\mathit{REVISE}}}
\newcommand{\Refute}{{\mathit{REFUTE}}}
\newcommand{\Reject}{{\mathit{REJECT}}}

\newcommand{\Init}{{\mathit{INIT}}}
\newcommand{\Match}{{\mathtt{MATCH}}}
\newcommand{\Agree}{{\mathtt{AGREE}}}
\newcommand{\magphys}{\textsc{magphys}}

\begin{document}

\title{Engineering Systems for Data Analysis Using Interactive Structured Inductive Programming}

\titlerunning{IS Engineering with Interactive Structured Programming}

\author{Shraddha Surana\inst{1} \and
Ashwin Srinivasan\inst{1} \and
Michael Bain\inst{2}}

\authorrunning{S. Surana et al.}

\institute{Dept. of Computer Science \& Information Systems, BITS Pilani, Goa, India\\
\email{p20220031@goa.bits-pilani.ac.in, ashwin@goa.bits-pilani.ac.in}
\and
School of Computer Science \& Engineering, University of New South Wales, Australia\\
\email{m.bain@unsw.edu.au}}

\maketitle              
%
\begin{abstract}

Engineering information systems for scientific data analysis presents significant challenges: complex workflows requiring exploration of large solution spaces, close collaboration with domain specialists, and the need for maintainable, interpretable implementations. Traditional manual development is time-consuming, while ``No Code'' approaches using large language models (LLMs) often produce unreliable systems. We present $\IP$, a tool implementing \textit{Interactive Structured Inductive Programming}. $\IP$ employs a variant of a `2-way Intelligibility' communication protocol to constrain collaborative system construction by a human and an LLM. Specifically, given a natural-language description of the overall data analysis task, $\IP$ uses an LLM to first identify an appropriate decomposition of the problem into a declarative representation, expressed as a Data Flow Diagram (DFD). In a second phase, $\IP$ then uses an LLM to generate code for each DFD process. In both stages, human feedback, mediated through the constructs provided by the communication protocol, is used to verify LLMs' outputs. We evaluate $\IP$ extensively on two published scientific collaborations (astrophysics and biochemistry), demonstrating that it is possible to identify appropriate system decompositions and construct end-to-end information systems with better performance, higher code quality, and order-of-magnitude faster development compared to Low Code/No Code alternatives. The tool is available at: \texttt{https://shraddhasurana.github.io/dhaani/}.


\end{abstract}

\section{Introduction}

Brooks and Dijkstra both noted the inherent complexity of software systems~\cite{Brooks1995,Dijkstra1972}. Despite advances in tools and abstractions, designing correct and maintainable software - especially for scientific data analysis - remains resistant to full automation. Scientific workflows involve data acquisition, transformation, modelling, validation, and reporting~\cite{Hey:etal:j:2020}, and must be built through sustained collaboration between software engineers and domain specialists. Traditional manual development is reliable but slow; purely LLM-driven ``No Code'' approaches are fast but often unreliable, producing code that lacks correctness, structure, or maintainability~\cite{Neja:etal:j:2025}.

LLMs trained on large codebases can generate substantial program fragments~\cite{fan2023largelanguagemodelssoftware,wu2024uicoderfinetuninglargelanguage,lu2024aiscientistfullyautomated}. Yet code generation alone is not software engineering: natural-language specifications are often ambiguous; integration issues compound; and LLMs struggle with multi-step reasoning and workflow design~\cite{Gu:etal:p:2025}. For complex analytical systems, a central bottleneck remains: determining an appropriate system decomposition before code can be generated.

This challenge echoes earlier work on \textit{structured induction}~\cite{shapiro1987:structInd,Bain2010:strucInd}, where a human decomposes a problem into sub-tasks and machine learning constructs sub-programs consistent with partial specifications. Modern LLMs allow us to revisit this idea using far more expressive program spaces, provided human oversight is preserved. In this paper, we present $\IP$, which implements \textit{Interactive Structured Inductive Programming} for information systems (IS) engineering using LLMs. Given a natural-language description of a scientific analysis task, $\IP$:

\begin{enumerate}
\item identifies a candidate workflow decomposition as a Data Flow Diagram (DFD) through LLM–human interaction;
\item generates code for each process using the ``2-way Intelligibility’’ protocol~\cite{2way}, where the human ratifies, refutes, or revises the LLM’s proposals; and
\item composes verified components into an end-to-end, executable system.
\end{enumerate}

\noindent
Our main contributions are:
\begin{itemize}
\item A methodology for semi-automated software engineering that uses an LLM to: (a) identify
a system decompositions from problem descriptions in natural language; and (b) generates human-ratified code, treating scientific data analysis as workflow-oriented IS development, following a communication
protocol specifically designed for intelligible exchange of messages;
\item Empirical evaluation on two published scientific collaborations demonstrating the
method is capable of identifying appropriate decompositions and construct an end-to-end
system with better performance than baseline Low Code/No Code alternatives.\footnote{We also compare
against a completely manual approach, but do not consider this an appropriate baseline for
reasons that will be clarified later.}
\end{itemize}

A subsidiary contribution of the work is the $\IP$ tool, which we see
as supporting the transition from natural-language problem descriptions to the development
of user-certified system development.

\section{Related Work}
\label{sec:relwork}

\paragraph{Information Systems Engineering.}
Recent low-code/no-code (LCNC) platforms~\cite{Rokis2024} accelerate IS development through visual programming and pre-built components. However, these platforms typically target standardized enterprise applications (e.g., CRUD operations, form-based workflows). They provide limited support for custom scientific data analysis pipelines requiring domain-specific transformations and complex analytical logic. Our approach addresses this gap by automating the engineering of non-routine analytical systems while maintaining the rigor of traditional IS methodologies~\cite{DeMarco2002,gane:strucsys,Wirth2000CRISPDMTA}.

\paragraph{LLMs for Code Generation.}
LLMs can generate executable code~\cite{fan2023largelanguagemodelssoftware,wu2024uicoderfinetuninglargelanguage} but face well-known limitations: ambiguity in natural-language specifications, inconsistent logical reasoning, and fragile integration across modules~\cite{Gu:etal:p:2025}. Empirical studies show substantial variation in correctness, including compilation and runtime errors~\cite{Neja:etal:j:2025}, and improvements only when structured human feedback is provided~\cite{austin2021}. Purely prompt-based workflow generation remains infeasible for complex IS tasks~\cite{enumprog2024}.
Assistive tools like Copilot~\cite{chen2021}, AutoML~\cite{hutter2011}, Cline~\cite{cline2025}, Replit~\cite{replit2025}, and AWS Kiro~\cite{kiro2025} support multi-file generation or ML automation but do not enforce architectural structure or verified module integration. 
Our ``No Code'' and ``Low Code'' baselines (Section~\ref{sect:expt}) approximate the interaction style of widely-used LLM assistants.
$\IP$ differs by combining: (a) explicit DFD-based workflow design, (b) semi-automated structure discovery, and (c) a human-controlled intelligibility protocol ensuring correctness and maintainability.

\paragraph{Structured and Inductive Programming.}
Structured induction~\cite{shapiro1987:structInd,Bain2010:strucInd} and earlier ML-based code generation~\cite{extran} recognised the value of decomposing programs into intelligible sub-tasks. Inductive Logic Programming (ILP) systems~\cite{Muggleton1987duce,Muggleton2023:deeplog,rocha2024}, Programming-by-Example (PBE)~\cite{gulwani2016flashfill}, and neuro-symbolic program synthesis~\cite{ellis2021} advance different aspects of program induction but typically require examples, DSLs (Domain Specific Language), or symbolic traces rather than natural-language workflow descriptions.
Structured prompting methods~\cite{wei:cot,hao2022structuredpromptingscalingincontext,Kramer2024UnlockingST,sahoo2024systematicsurveypromptengineering} improve LLM reasoning by constraining prompts. Our approach generalises this idea: descriptions, pre/post-conditions, and DFDs collectively act as structured prompts guiding LLM behaviour, while the interaction protocol ensures human-verifiable outputs.

\paragraph{Interaction Models.}
Our interaction adapts the 2-way Intelligibility protocol~\cite{2way}, which formalizes when agent interactions are mutually intelligible via message tags. The studies here demonstrate practical utility, addressing a gap in~\cite{2way}. The protocol-based approach enables formal reasoning about interaction properties, distinguishing it from open-ended dialogue models~\cite{madumal}.

\section{iProg: Interactive Structured Inductive Programming}
\label{sect:strucind}

We present $\IP$'s approach to semi-automated IS engineering, starting with structure learning and then code generation.
\paragraph{Inductive Programming:}
Inductive programming is commonly viewed as identifying a program that satisfies a partial specification, given background knowledge and a search bias over candidate programs. In our setting, the ``program'' is not a single function but an information system: a workflow of connected components that must be correct both locally (each component satisfies its specification) and globally (interfaces match and the end-to-end system executes). This motivates a structured approach in which (i) the system’s decomposition is made explicit (so that correctness and maintainability can be assessed at the module level), and (ii) code synthesis is performed component-wise under a verification-oriented interaction protocol.

\noindent
We assume the following informal specification of Inductive Programming (IP).

\smallskip \noindent
\textbf{Given:}
(a)  Prior or background knowledge, describing domain-specific and  domain-independent information possibly relevant to the program identification process;
(b) A specification of the program to be identified; and (optionally)
(c) functions and relations for enumerating
    elements from the (possibly infinite) set of programs; \\
\textbf{Find:} A program that is consistent with the specifications.

\smallskip \noindent
Reviews of IP can be found in~\cite{schmid2006survey,gulwani:ip}. We extend IP to systems engineering by first identifying system decomposition (structure), then constructing component programs.

\paragraph{Structured induction as a two-stage process:} Following the perspective of structured induction, system construction can be viewed as two coupled stages: (1) top-down structure identification, where the problem is decomposed into a solution structure (here, a DFD capturing processes and data flows); and (2) bottom-up component synthesis, where executable programs are constructed for each sub-task and composed into an end-to-end system. $\IP$ operationalises this view for IS engineering: $\IS$ performs interactive structure identification from natural language, and $\IC$ performs interactive component synthesis and integration from the resulting DFD.
\subsection{Interactive Structure Identification} \label{sec:structure}

The first step in $\IP$ is identifying an appropriate system decomposition from a natural language problem description. We represent decompositions as Data Flow Diagrams (DFDs)~\cite{DeMarco2002,gane:strucsys}, a foundational IS design technique providing declarative workflow representation.\footnote{Workflow models such as CRISP-DM provide a high-level decomposition of data mining projects into phases (e.g., business understanding, data understanding, preparation, modelling, evaluation, deployment). In $\IP$, DFD process nodes represent operationalised sub-tasks (guarded functions with concrete pre-/post-conditions) and therefore correspond more closely to lower-level activities within or across such phases, rather than mapping 1:1 to CRISP-DM phases. We use CRISP-DM only to motivate the general premise that data analysis work is naturally workflow-shaped.}

DFDs are directed acyclic graphs with vertices representing processes (system components), data sources or data stores, and edges representing information flow. Process vertices are labeled with:
\begin{itemize}
\item \textbf{Description}: Natural language specification of the sub-task
\item \textbf{Pre-condition}: Required state/data for execution
\item \textbf{Post-condition}: Guaranteed state/data after execution
\end{itemize}

\begin{myexample}[Semi-Automated DFD Identification]
\label{ex:dfd_learning}
Consider the following natural language problem description for a biochemistry analysis:

\textit{``Classify protein sequences as antimicrobial peptides (AMPs) or non-AMPs using machine learning with alphabet reduction and distributed vector representations. Use Uniprot data to create embedding models for combinations of reduction techniques, k-mer values, context windows, and vector sizes. Create classification dataset from public databases. Train and evaluate models.''}

Given this description, $\IP$ proposes an initial DFD with processes such as: (P1) Download Uniprot data; (P2) Apply alphabet reduction; (P3) Create embedding models; (P4) Load classification datasets; (P5) Create sequence embeddings; (P6) Train classifiers; (P7) Evaluate models. The software engineer reviews this proposal, refines granularity, clarifies interfaces, and ratifies the final structure (see Figure~\ref{fig:dfd_task2} in Section~\ref{sec:results}).
\end{myexample}

Natural-language specifications are often ambiguous or underspecified, and small interface misunderstandings can compound across workflow modules. $\IP$ mitigates this by making assumptions explicit via DFD structure together with pre-/post-conditions, and by using RATIFY/REFUTE/REJECT/REVISE tags to force convergence.

Procedure~\ref{alg:is} implements $\IS$ for interactive structure identification. It adapts the 2-way Intelligibility protocol~\cite{2way} to DFD proposals: the LLM proposes DFDs and explanations; the human responds with tags ($\Ratify$, $\Refute$, $\Reject$) based on whether the proposed DFD and explanation are acceptable. Interaction terminates when the human ratifies a DFD or rejects after multiple unsuccessful attempts.

{\small{
\begin{algorithm}[!htb]
    \caption{$\IS$}
    \label{alg:is}
    \textbf{Input}: 
        $\mathcal{T}$: description of the problem;
        $\lambda$: an LLM-based agent;
        $h$: a human-agent;
        $R$: an upper-bound on the number of retries to obtain a non-empty DFD;
        $n$: upper-bound on number of messages exchanged between $\lambda$ and $h$;
        $m$: message-number after which an agent can send $\Reject$ tags in messages; $k$: max attempts\\
    \textbf{Output}: $\Sigma$: a  DFD decomposition of $\mathcal{T}$.

    \begin{algorithmic}[1]

    \State $\Sigma := \Box$\;      {\scriptsize{// Empty DFD}}
    \State $C_0 := \emptyset$\;  {\scriptsize{// Initial context for the LLM}}
    \State $i := 1$\;
    \State $Failed:= FALSE$\;
    \While{$((\neg Failed) \wedge (i \leq k))$} 
        \State ${Spec}_i := \mathcal{T}$\;
        \State $(\Sigma_i,C_i) =$ $\INTSTRUC(\lambda,h,{Spec}_i,C_{i-1},R,n,m)$\; {\scriptsize{// Construct DFD interactively with the LLM}}
        \State $Failed:= (\Sigma_i = \Box)$\;
        \State increment $i$\;
    \EndWhile
    \If{$((i > k)$}
        $\Sigma := \Sigma_{i-1}$\;
    \EndIf\\
    \Return{$\Sigma$}
    \end{algorithmic}
\end{algorithm}
}}

The key procedure $\INTSTRUC$ is very similar to $\INTCODE$. 
Procedure $\INTSTRUC$ returns a user-ratified DFD and the context (or $(\Box,\emptyset)$).

\subsection{Interaction Model}
\label{ref:icode}

Given a ratified DFD from $\IS$, $\IC$ generates code for each process. We base human-LLM interaction on the $\mathtt{PEX}$ protocol~\cite{2way}, which addresses when interactions between agents are \textit{intelligible}. This is critical for ensuring the generated code is comprehensible. Key features of the interaction are:

\begin{enumerate}[(i)]
\item Messages between agents carry tags: $\Ratify$, $\Refute$, $\Revise$, or $\Reject$;
\item Interaction begins with the human sending $(\Init,Spec,?)$ where $Spec = (Descr,Pre,Post)$ is a process specification (see Example~\ref{ex:process});
\item Machine (LLM) responds with $(T_m,P_m,E_m)$ where $T_m \in \{\Ratify, \Revise, 
\newline \Refute\}$, $P_m$ is generated code, $E_m$ is an explanation;
\item Human responds with $(T_h,P_m,E_h)$ where $T_h \in \{\Ratify, \Refute, 
\newline \Reject\}$; $P_m$ is the code received from the machine; and $E_h$ is the human's explanation.
\end{enumerate}

Tags are decided by functions $\Match$ and $\Agree$ defined over programs and explanations, as shown in Figure~\ref{fig:tags}. The interaction is asymmetric: humans never send $\Revise$; machines never send $\Reject$. In practice, we do not require either agent to have an explicit encoding of functions for
 $\Match$ and $\Agree$, but assume the agent is able to decide on their truth-values.
 That is, the machine-agent sends a $\Revise$ tag if it either changes its program or its explanation;
 the human-agent sends a $\Refute$ tag if the software engineer does not find one of the
 program or explanation to be acceptable; and so on.

\begin{figure}[t!]
\centering
\begin{subfigure}[t]{0.48\textwidth}
{\small{
\begin{center}
\begin{tabular}{cc|c|c|}
      &\multicolumn{1}{c}{\mbox{}} & \multicolumn{2}{c}{Explanation} \\
  \multirow{3}{*}{\parbox[c]{0.5cm}{\centering \rotatebox{90}{Program      }}}   & \multicolumn{1}{c}{\mbox{}} & \multicolumn{1}{c}{$\Agree_h$}    & \multicolumn{1}{c}{$\neg \Agree_h$}  \\ \cline{3-4}
            & $\Match_h$        & $\Ratify$ & $\Refute$      \\
            &              &           &                \\ \cline{3-4}
            &  $\neg \Match_h$ & $\Refute$ & $\Refute$ or   \\
            &              &           & $\Reject$      \\ \cline{3-4}
\end{tabular}
\end{center}
}}
\caption{Human}
\end{subfigure}%
\hfill
\begin{subfigure}[t]{0.48\textwidth}
\centering
{\small{
\begin{center}
\begin{tabular}{cc|c|c|}
&\multicolumn{1}{c}{\mbox{}}&\multicolumn{2}{c}{Explanation} \\
\multirow{3}{*}{\parbox[c]{0.5cm}{\centering \rotatebox{90}{Program    }}}  &\multicolumn{1}{c}{\mbox{}}&\multicolumn{1}{c}{$\Agree_m$}&\multicolumn{1}{c}{$\neg \Agree_m$} \\   \cline{3-4}
            & $\Match_m$         & $\Ratify$     & $\Refute$ or  \\
            &               &               & $\Revise$     \\ \cline{3-4}
            &  $\neg \Match_m$  & $\Refute$ or  & $\Refute$     \\
            &               & $\Revise$     &               \\ \cline{3-4}
\end{tabular}
\end{center}
}}
\caption{Machine}
\end{subfigure}
\caption{Deciding tags for messages during code identification.}
\label{fig:tags}
\end{figure}
\begin{myexample}[Process Element in a DFD]
\label{ex:process}
Sub-tasks for a problem are represented by process elements in the DFD.
Each process element is to be understood as implementing a function that accomplishes
the sub-task. One such sub-task in the BIO problem is
to create embedding models for sequence data.
This is represented by a process element containing the following.

\noindent
\textbf{Description:} Use Word2Vec to create embeddings of protein sequences for various combinations
of k-mer: 3 and 5; context window: 5, 10 
and 25; vector size: 100, 200 and 300; and 
training model: skip-gram.
    
\noindent
\textbf{Pre-condition:} The Swiss-Prot and reduced 
sequences files are present in the 
'reduced\_sequences' directory.
    
\noindent
\textbf{Post-condition:} Embedding models created 
and saved for later retrieval.
\end{myexample}

\noindent
Computationally, the sub-function represented by a process element in a DFD
will be implemented by a corresponding (sub-)program. We first consider the
problem of obtaining a program for a single process-element
containing a natural-language specification of the kind shown in Example~\ref{ex:process}.
Specifically, we provide an interactive procedure by which a software engineer communicating with an LLM can obtain human-certified code.
Procedure~\ref{alg:interact} implements $\INTCODE$ for interactive code identification.



{\small{
\begin{algorithm}[!htb]
    \caption{$\INTCODE$}
    \label{alg:interact}
    \textbf{Input}: $\lambda$: an LLM-based agent;
        $h$: a human agent;
        $Spec$: a specification consisting of the function description, pre-condition, and post-condition;
        $C$: an initial context for the LLM;
        $R$: an upper-bound on the number of retries to obtain a non-empty program;
        $n$: an upper-bound on the number of messages exchanged between $\lambda$ and $h$;
        $m$: message-number after which an agent can send $\Reject$ tags in messages\\
    \textbf{Output}: $(P,C)$ where $P$ is a program and $C$ is a context.

    \begin{algorithmic}[1]
    \State Let $Spec = (F,Pre,Post)$\;
    \State $P := \Box$\;
    \State $C_0 := C$\;
    \State $r := 1$\;
    \State $Done := FALSE$\;
    \While{$(\neg Done) \wedge (r \leq R)$}
        \State $P_0 := \Box$\;
        \State $\mu_0 := (\Init,?,?)$\;
        \State $C_r := C_{r-1}$\;  {\scriptsize{//Retain context from previous retry}}   
        \State $i := 1$\;
        \State $Found := FALSE$\;
        \While{$(\neg Found) \wedge (i \leq n))$}
            \State $C_r := C_{r} \cup \{\mu_{i-1}\}$\; {\scriptsize{//Update context with message sent}}
            \State $({Tag}_\lambda,P_\lambda,E_\lambda) := GenCode(\lambda,Spec,C_r)$\; \label{step:llm} {\scriptsize{// Get a program and explanation}}
            \State ${Tag}_h := \mathit{EVAL}_h(P_\lambda,E_\lambda)$\; {\scriptsize{//Decide tag based on correctness of $P_\lambda$, $E_\lambda$}}
            \If {$({Tag}_h := \Reject) \wedge (i \leq m)$}
                \State ${Tag}_h = \Refute$\;  {\scriptsize{//Only start rejecting after $m$ steps}}
            \EndIf
            \State $Found := (({Tag}_h = \Ratify) \vee ({Tag}_h = \Reject))$\;
            \If{$Found$}
                \State $P_i := P_\lambda$\;
            \Else
                \State $P_i = \Box$\;
                \State Let $E_h$ be a refutation of $(P_\lambda,E_\lambda)$\;
                \State $\mu_i := ({Tag}_h,P_\lambda,E_h)$\;\label{step:refute}
           \EndIf
            \State increment $i$\;
        \EndWhile
        \State $Done := (P_{i-1} \neq \Box)$\;
        \State $P = P_{i-1}$\;
        \State increment $r$\;
    \EndWhile \\
    \Return{$(P,C_{r-1})$}
\end{algorithmic}
\end{algorithm}
}}



\subsection{End-to-end system construction}

$\IP$ orchestrates structure learning and code generation to construct complete information systems. The tool first invokes $\IS$ (Procedure~\ref{alg:is}) to obtain a ratified DFD (which can be
viewed as the background knowledge provided for inductive programming), then uses the $\IC$ procedure (Procedure~\ref{alg:ic}) to traverse the DFD (by examining the total-orderings of processes), invoking $\INTCODE$ for each process. Ratified sub-programs are composed according to DFD connectivity\footnote{For clarity, the implementation of $\IC$ shown
does not contain measures taken for efficiency. Thus, as shown, $\IP$ may
repeatedly construct a program for the same function as we explore different
orderings. This is clearly not required; once we have obtained an acceptable program for a function, it is stored for re-use.}.

{\small{
\begin{algorithm}[!htb]
    \caption{$\IC$}
    \label{alg:ic}
    \textbf{Input}: 
        $\mathcal{T}$: description of the problem;
        $\Sigma$: a DFD representing a decomposition of the problem;
        $\lambda$: an LLM-based agent;
        $h$: a human-agent;
        $R$: an upper-bound on the number of retries to obtain a non-empty program;
        $n$: upper-bound on number of messages exchanged between $\lambda$ and $h$;
        $m$: message-number after which an agent can send $\Reject$ tags in messages\\
    \textbf{Output}: $(P,\phi)$, where $P$ is a set of programs implementing
        the functions in $\Sigma$, and $\phi$ is an end-to-end program for the problem
        described by $\mathcal{T}$.

    \begin{algorithmic}[1]
    \State $P := \emptyset$\;
    \State $\phi := \Box$\;      {\scriptsize{// Empty program}}
    \State Let $(v_1,v_2,\ldots,v_k)$ be an ordering of vertices consistent with a breadth-first traversal
        of $\Sigma$\;
    \State $C_0 := \mathcal{T}$\;  {\scriptsize{// Initial context for the LLM containing the task description}}
    \State $i := 1$\;
    \State $Failed:= (\Sigma = \Box)$\;
    \While{$((\neg Failed) \wedge (i \leq k))$} 
        \State Let ${Spec}_i =  ({Descr}_i,{Pre}_i,{Post}_i)$ be the vertex-label of $v_i$\;
        \State $(P_i,C_i) =$ $\INTCODE(\lambda,h,{Spec}_i,C_{i-1},R,n,m)$\; {\scriptsize{// Construct program interactively with the LLM}}
        \State $Failed:= (P_i = \Box)$\;
        \State increment $i$\;
    \EndWhile
    \If{$((i > k)$}
        $P := \{P_1,P_2,\ldots,P_k\}$\;
        $\phi := P_1 \oplus P_2 \cdots \oplus P_k$\; {\scriptsize{// Concatenation of sub-programs}}

    \EndIf\\
    \Return{$(P,\phi)$}
    \end{algorithmic}
\end{algorithm}
}}

\noindent
The $\IC$ procedure traverses the DFD in breadth-first order, generating code for each process while accumulating context. The final system $\phi$ is a composition of validated components.

\subsection{The $\IP$ Tool}
\label{sec:tool}
It is evident that the DFD identified in Procedure \ref{alg:is} can be provided
as input to Procedure \ref{alg:ic}. That is, Structured Inductive Programming can
be viewed as the generalised composition of the functions implemented by $\IC$ and $\IS$. 
To allow the exploration of this form of program identification, as well as (optionally) program
identification with a user-defined decomposition, we have developed the $\IP$ tool.
At present, $\IP$ has the following functionalities:
(a) Accepting a user-defined DFD, or interactive identification of a DFD from a problem-specification
in natural language; 
(b) Manual editing of a DFD through a graphical user-interface;
(c) Interactive identification of sub-programs associated with process-elements of a DFD and end-to-end
    code for a problem; and
(d) Facilities to run code for individual sub-programs. Fig.~\ref{fig:dfd_task2} shows a DFD created using the $\IP$ tool.
The tool is available at: \texttt{https://shraddhasurana.github.io/dhaani/}.


\section{Empirical Evaluation}
\label{sect:expt}

\subsection{Aim(s)}
\label{sec:aims}

Our principal aim is to investigate the use of $\IP$ to
construct information systems for scientific data analysis. Specifically, using two non-trivial scientific data analysis tasks, we test whether:
\begin{enumerate}[(a)]
\item $\IP$ can identify appropriate system decompositions from problem descriptions;
\item systems constructed by $\IP$ can perform at least as well as Low Code/No Code alternatives along dimensions of performance, software quality, and development effort;
\item $\IC$ can construct a program that correctly achieves the post-conditions specified for each problem; and
\item as a secondary question, how $\IP$-constructed systems compare against manually developed systems.
\end{enumerate}

\subsection{Materials} 
\label{sec:mat}

\subsubsection*{Problems}

We examine two scientific data analysis information systems developed through Thoughtworks Inc.'s ``Engineering for Research'' (E4R) initiative, originally constructed by multi-engineer teams collaborating with domain scientists, with results published in peer-reviewed venues.

\noindent
\textbf{Astrophysics Information System (PHY).}
Predicts star formation properties (rate, dust luminosity, stellar mass) from multi-spectral astronomical data (21 bands, 76,455 galaxies from GAMA catalogue~\cite{Driver2011:gama}). Workflow: data loading, exploratory analysis, feature engineering, ML model training, visualization. Published in Monthly Notices of the Royal Astronomical Society (MNRAS)~\cite{sfp}. Manual development error rates (test data): star formation 0.164, dust luminosity 0.114, stellar mass 0.058.

\noindent
\textbf{Biochemistry Information System (BIO).}
Classifies protein sequences as antimicrobial peptides using ML with alphabet reduction and distributed vector representation. Uses Uniprot~\cite{uniprot} for embedding models (combinations of reduction techniques, k-mer, context window, vector size); 14,821 sequences from public databases. Workflow: data acquisition, alphabet reduction, embedding creation, classification, evaluation. The outcomes of this study were published in the proceedings of the IEEE International Conference on Bioinformatics and Biomedicine (BIBM: \cite{amp_classification}). Manual development error: 0.021.


We refer the reader to the original studies for further details of the problems, terminology, and approaches used.

\subsubsection{Evaluation Design}

For both systems, we provide $\IP$ with only the natural language problem description (no DFD, no code). We measure:

\textbf{Structure Quality}: Compare DFDs identified by $\IP$ against DFDs used in manual development along vertices (process count), edges (data flow connections), and label agreement (engineer-judged semantic equivalence).

\textbf{System Performance}: Error on test data (PHY: $error = \sigma(y_{actual}-y_{predicted})$ where $y_{actual}$ is \magphys{}~\cite{daCunha2008} value; BIO: $error = 1-accuracy$).

\textbf{Code Quality}: Logic checks (pylint~\cite{pylint}); type violations (mypy~\cite{mypy}); cyclomatic complexity (radon~\cite{radon}); lines of code.

\textbf{Development Effort}: Days to completion; number of engineers; person-months; human-machine interactions.

\textbf{Baselines}: No Code (LLM-0, CP-0: direct generation); Low Code (LLM-k, CP-k: generation with free-form human feedback, interaction count matched to $\IP$); Manual (original published systems). For CP-k and LLM-k, the same engineer provided unconstrained natural-language feedback using similar efforts as with $\IP$, but without DFD-based structuring or protocol tags.

\paragraph{Artifacts and reproducibility:} Artifacts for repeatability (problem descriptions, final DFDs, per-process specs, and generated code) are available as public runnable demo projects in the Dhaani-iProg tool.

\subsection{Method} \label{sec:meth}

\textbf{Procedure}:
\begin{enumerate}
\item Provide $\IP$ with the problem description
\item $\IP$ invokes $\IS$ to learn DFD (engineer validates/refines proposals)
\item Obtain the overall program using $\IC$ which in turn calls $\INTCODE$ for each process to generate code
\item Compose and test resulting system
\item Compare against baselines
\end{enumerate}

\textbf{Parameters} (empirically determined): $R = 5$, $n = 10$, $m = 6$. LLM temperature: $1.0$.
\textbf{LLM}: Azure OpenAI GPT-4~\cite{openai2024gpt4technicalreport}.
\textbf{Hardware}: 2.4 GHz Intel Core i9, 64 GB RAM; Ubuntu 24.04, AMD Ryzen 9 5900X, 64 GB RAM.


\subsection{Results}
\label{sec:results}

Figure~\ref{fig:results_is} presents structure identification results. $\IP$ successfully identified DFDs for both problems with high agreement with manual decompositions.

\begin{figure}[htb]
    \centering
    {\small{
    \begin{tabular}{|l|c|c|c|c|c|} \hline
        Problem & Manual  & $\IP$   & $I$ & \multicolumn{2}{|c|}{Agreement} \\ \cline{5-6}
                & $(V,E)$ & $(V,E)$ &     &  $(V)$ & $(E)$ \\ \hline
        PHY     & (4, 4)  & (4, 6)  & 3 & 4 & 4\\  \hline
        BIO     & (8, 10) & (8, 12)  & 4 & 8 & 12  \\ \hline
    \end{tabular}
    }}
    \caption{DFD structure learning results. Manual: original development DFD (ground-truth). $V,E$: process vertices and edges; $I$: interactions for ratified DFD; Agreement: matching vertex/edge labels (engineer-judged).}
    \label{fig:results_is}
\end{figure}

\begin{figure}[!htb]
    \centering
    \includegraphics[width=1.0\linewidth]{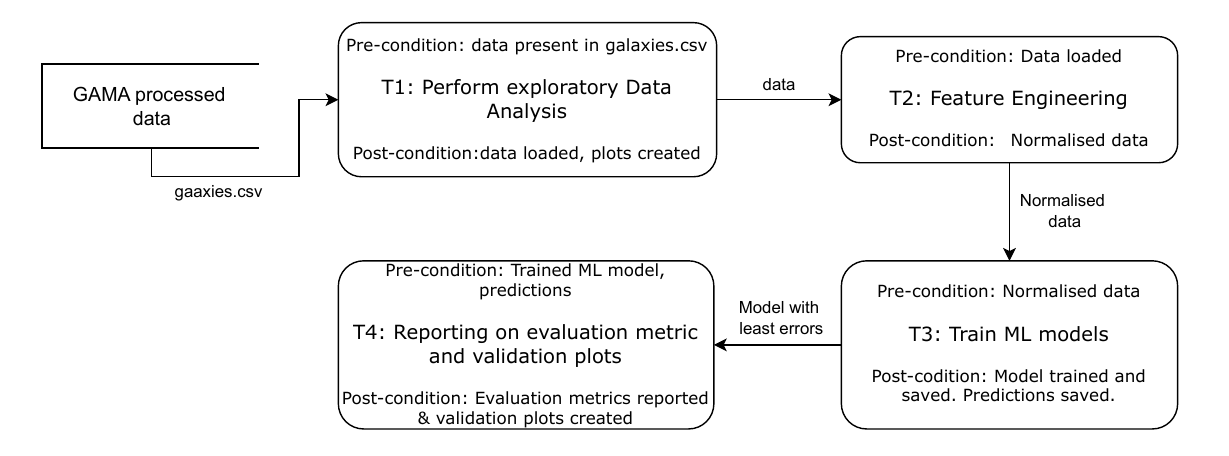}
    \caption{DFD for PHY system learned by $\IP$ from problem description. Manually drawn to show the pre and post conditions of each process.}
    \label{fig:dfd_task1}
\end{figure}

\begin{figure*}[!htb]
    \centering
    \includegraphics[width=0.95\linewidth]{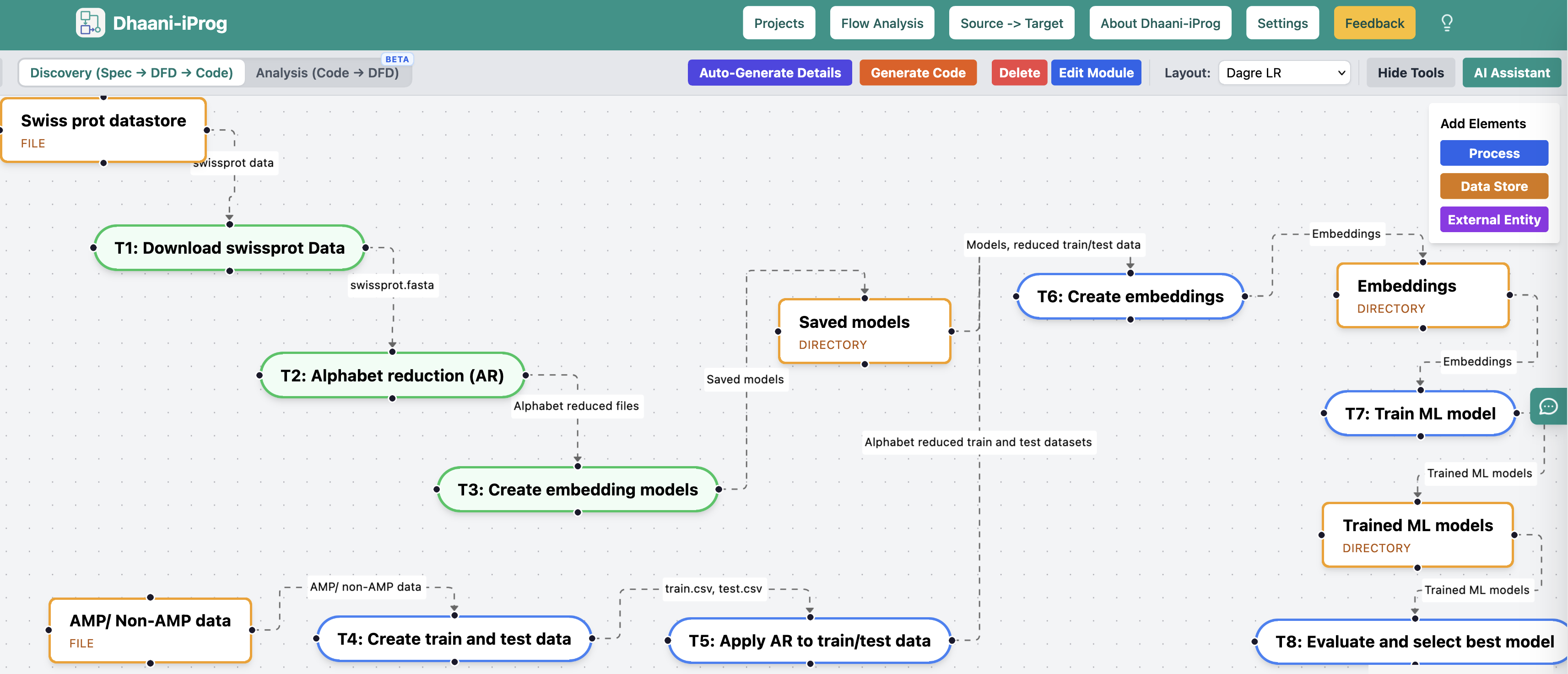}
    \caption{DFD for BIO system learned by $\IP$ from problem description. The specification, pre and post conditions are hidden in the DFD view and are visible within each process page.}
    \label{fig:dfd_task2}
\end{figure*}

For PHY, $\IP$ identified 4 processes matching manual decomposition. Minor edge differences arose from $\IP$ adding explicit initialization edges. 
For BIO, $\IP$ initially created finer-grained decomposition (separate data loading and train-test split vs. combined), creating more processes. Through interactive refinement, the final DFD created had the same number of processes but more data stores - which we recognized as a better workflow that saves intermediate artifacts rather than regenerating them during each end-to-end execution.
The DFDs learned by $\IP$ represent appropriate, workable decompositions. The final DFDs ratified by the user are shown in Fig.~\ref{fig:dfd_task1} and Fig.~\ref{fig:dfd_task2} for the PHY and BIO problems respectively.


Figure~\ref{fig:results1} presents system performance, code quality, and development effort comparing $\IP$ against LCNC baselines. Errors from $\IP$ are consistently lower than LCNC approaches. ``No Code'' approaches (CP-0, LLM-0) frequently fail with runtime errors or produce non-functional systems. In such cases (marked ``Error'' or ``--''), downstream quality metrics are not applicable because no executable end-to-end system is produced; we therefore report code-quality measures only for functional systems. ``Low Code'' approaches (CP-k, LLM-k) with unstructured human feedback achieve some success on PHY but still exhibit substantially higher errors than $\IP$, and fail completely on the more complex BIO problem.

Code quality metrics reveal substantial differences. For PHY, $\IP$ achieves logic scores of 6.69/10 compared to 1.71--1.88 for Low Code baselines. The lower line count in Low Code implementations (48--50 versus $\IP$'s 259) reflects their monolithic structure rather than true simplicity. In contrast, $\IP$ generates structured, modular code with explicit interfaces between components. The very low programming effort by Low Code/No Code options is unsurprising given their minimal interaction, whereas $\IP$ attains substantially better performance with a similar number of interactions. For BIO, Low Code approaches fail to produce functional systems entirely, preventing quality assessment.

The results with LLM-k provide evidence for the combined effect of structuring and the interaction protocol. The LLM used in both $\IP$ and LLM-k is the same, as are the number of interactions. What differs is that LLM-k does not employ explicit DFD-based structuring, and feedback does not follow the protocol's specification (tags, pre/post-conditions)--resulting in substantially worse outcomes.

\begin{figure}[htb]
\begin{subfigure}{\linewidth}
    \centering
    {\tiny{
    \begin{tabular}{|l|l|c||c|c|c|c|}
        \hline
        Problem & Prediction & $\IP$ & \multicolumn{2}{|c|}{Low Code} & \multicolumn{2}{|c|}{No Code} \\ \cline{4-7}
                &            &       & CP-k   &   LLM-k & CP-0    & LLM-0 \\ \hline
                & Star formation   & 0.030 & 0.332    & 0.357 & Error & Error\\
        PHY     & Dust luminosity  & 0.022 & 0.133   & 0.195 & Error   & Error\\
                & Stellar mass     & 0.020 & 0.140    & 0.149 & Error   &Error\\
        \hline
        BIO     & AMP Classification & 0.024 & Error   & Error   & Error & Error \\
        \hline
    \end{tabular}
    }}
    \subcaption{System Performance (Error on test data)}
\end{subfigure}

\begin{subfigure}{\linewidth}
\centering
    {\tiny{
    \begin{tabular}{|l|l|c||c|c|c|c|}
        \hline
        Problem & Measure           & $\IP$  & CP-k      & LLM-k & CP-0  & LLM-0\\ \hline
                & Logic score (/10) & 6.69   & 1.88      & 1.71  & --    & -- \\
         PHY    & Type errors       & 1      & 0         & 0     & --    & -- \\
                & Complexity        & 1.11   & 1.33      & 1.0   & --    & -- \\
                & Lines of code     & 259    & 48        & 50    & --    & -- \\ \hline
                & Logic score (/10) & 6.32   &  --       & --    & --    & -- \\
         BIO    & Type errors       & 0      & --        & --    & --    & -- \\
                & Complexity        & 2.17   & --        & --    & --    & -- \\
                & Lines of code     & 754    &  --       & --    & --    & --\\ \hline
    \end{tabular}
    }}
    \subcaption{Code Quality (-- indicates non-functional system)}
\end{subfigure}

\begin{subfigure}{\linewidth}
    \centering
    {\tiny{
    \begin{tabular}{|l|l|c||c|c|c|c|}
    \hline
        Problem &  Measure   & $\IP$ & CP-k  & LLM-k   & CP-0  & LLM-0\\ \hline
                & Days (approx)     & 4  & 4         & 4     & 1     & 1\\
        PHY     & People            & 1  & 1         & 1     & 1     & 1\\
                & PMs (approx)      & 0.13  & 0.13      & 0.13  & 0.03  & 0.03\\
                & Interactions      & 16 (3 struct + 13 code) & 13  & 13    &1  & 1\\
        \hline
                & Days (approx)     & 10 & 5         & 4     & 1     & 1 \\
        BIO     & People            & 1  & 1         & 1     & 1     & 1 \\
                & PMs (approx)      & 0.3 & 0.2      & 0.13  & 0.03  & 0.03\\
                & Interactions      & 26 (4 struct + 22 code) & 22  & 22    & 1  & 1\\
        \hline
    \end{tabular}
    }}
    \subcaption{Development Effort}
\end{subfigure}
\caption{Comparison of $\IP$ (semi-automated structure learning + code generation) against LCNC alternatives.}
    \label{fig:results1}
\end{figure}

\textbf{Comparison with Manual Development}: While a detailed comparison with manual development involves potential bias (the first author here participated in the original development).
Nevertheless, the published manual systems provide some reference points:
PHY manual errors were 0.058--0.164 versus $\IP$'s 0.020--0.030; BIO manual error was 0.021 versus $\IP$'s 0.024 (comparable). Manual development required 2--3 engineers over 30--60 days (2--6 person-months), versus $\IP$'s single engineer over 4--10 days (0.13--0.3 person-months).

Taken together, the results tabulated here suggest that rapid,
high-quality scientific data analysis programs can be constructed by
a software engineer interacting with a commodity LLM using
interactive structured induction implemented by $\IP$.  Further, there is evidence that constructing
programs for data analysis in this manner may be better than Low Code/No Code options.\footnote{
We have also verified that  $\IC$ and $\IS$ functionality within $\IP$ work as expected for the
problems in \cite{shraddha_iparc}. This indicates that the $\IP$ tool has a wider
applicability than just the data-analysis problems considered here.
}

\textbf{Summary}: $\IP$ successfully learns appropriate system decompositions from problem descriptions and constructs high-quality information systems, outperforming manual development and LCNC alternatives along performance, quality, and effort dimensions.

\smallskip\noindent
Our study is based on two scientific data analysis systems and mainly uses one LLM configuration, with the first author involved in both manual and $\IP$-based developments. 
Our evaluation was conducted by a single experienced software engineer; therefore, some expert judgment (e.g., problem decomposition and resolving ambiguity in natural-language requirements) is implicit, and assessing usability with less-experienced engineers or domain experts is left to future work. Sanity checks with other LLMs (e.g., Claude, Gemini and Ollama) yielded different DFDs and code but confirmed that the underlying interaction and decomposition concepts are model-independent. We mitigate potential bias by keeping evaluation procedures consistent across $\IP$ and LCNC baselines (including CP/LLM variants) and by having previously validated the $\IC$ and $\IS$ components on the IPARC problems~\cite{shraddha_iparc}, but further studies in other IS domains and with additional models would strengthen the evidence.

\subsection{Key Takeaways}

Our case studies yield several practical insights about human-LLM collaboration for information systems engineering:
\begin{enumerate}
    \item \textbf{Structuring enables understanding}: The DFD-based modular structure facilitated a better understanding of the problems and targeted modifications. When adjustments were
    needed (e.g., changing embedding parameters in BIO), only the relevant process required re-generation, leaving other validated components unchanged. This contrasts with monolithic approaches, where local changes risk system-wide impacts.
     The declarative specification that results (descriptions, pre/post-conditions) enabled clearer communication between human and LLM by reducing
     ambiguity inherent in free-form natural language feedback.
    
    \item \textbf{LLMs can propose meaningful structuring}: Given high-level problem descriptions, the LLM was usually able to organise data analysis tasks into DFDs that aligned with how domain experts thought about the workflow (e.g., separate stages for data preparation, modelling, and evaluation). These proposals were rarely perfect in the first attempt, but they provided a useful starting point that the human could refine and ratify with a few interactions.
    
    \item \textbf{Protocol-driven interaction is useful and practical}: The protocol-based interaction with
    explicit tags ($\Ratify$, $\Refute$, $\Reject$) allowed
    convergence in 3-4 interactions for identifying the structure (both problems),
    and code generation required
    13--22 overall (about 2--3 per process). These short, focused exchanges suggest the approach scales to real-world IS engineering without excessive human effort.

   \item \textbf{Human-verification is needed}: In both case studies, a human was needed to ratify or reject LLM proposals, for both structure and code. With this verification in place, we could converge to acceptable DFDs and working implementations in a small number of iterations; without it, errors and inconsistencies persisted and were hard to detect.
    LCNC-style approaches that attempted to generate and execute end-to-end pipelines without explicit human checks frequently produced brittle or opaque solutions. Our results indicate that such tools are better seen as components in an interactive process, rather than as autonomous replacements for human-guided design, implementation and validation.
\end{enumerate}

\section{Conclusions}
\label{sec:conc}

We have presented $\IP$, a tool implementing \textit{Interactive Structured Inductive Programming} for semi-automated information systems engineering. Given natural language problem descriptions, $\IP$ automatically learns appropriate workflow decompositions and generates human-ratified code for each component through structured human-LLM interaction.

The approach can be enhanced in several directions. More sophisticated search procedures beyond breadth-first traversal could improve efficiency, and supporting cyclic DFDs would enable workflows with feedback loops. Further refinement of structure learning and context management (e.g., hierarchical DFDs, summarised interaction histories) could potentially enable domain specialists to construct and evolve systems more directly.

Automated information systems engineering through interactive structured inductive programming offers a promising path forward. By leveraging LLM knowledge for initial proposals while maintaining human validation, we can accelerate IS construction while ensuring quality, maintainability, and correctness. While our evaluation focuses on scientific data analysis, the techniques are applicable across diverse domains requiring workflow-oriented information systems, including energy analytics, environmental monitoring, and other data-intensive fields~\cite{AIforEnergy2024}.

\bibliographystyle{splncs04}
\bibliography{bibliography}

%
%
%
%

\end{document}